\documentclass[final]{clv2025}

\jvol{vv}
\jnum{nn}
\jyear{2025}
\dochead{Long Paper} 
\pageonefooter{Action editor: \{action editor name\}. Submission received: DD Month YYYY; revised version received: DD Month YYYY; accepted for publication: DD Month YYYY.}

\usepackage{amsmath, amssymb}
\usepackage{graphicx}
\usepackage{booktabs}
\usepackage{subcaption} 
\usepackage{array}
\usepackage{hyperref}
\usepackage{float}
\usepackage{multirow} 

\runningtitle{Contextual Gating within the Transformer Stack for Classification}
\runningauthor{Gameiro}

\begin{document}

\title{\textbf{Contextual Gating within the Transformer Stack: Synergistic Feature Modulation for Enhanced Lyrical Classification and Calibration}}

\author{Marco Gameiro\thanks{Corresponding author}}

\affilblock{
    \affil{Independent Researcher, Amsterdam, Netherlands\ \quad \email{marcogameir@hotmail.com}}
}

\maketitle

\begin{abstract}
This study introduces a significant architectural advancement in feature fusion for lyrical content classification by integrating auxiliary structural features directly into the self-attention mechanism of a pre-trained Transformer. I propose the SFL Transformer, a novel deep learning model that utilizes a Contextual Gating mechanism (an Intermediate SFL) to modulate the sequence of hidden states within the BERT encoder stack, rather than fusing features at the final output layer. This approach modulates the deep, contextualized semantic features ($H_{\text{seq}}$) using low-dimensional structural cues ($F_{\text{struct}}$). The model is applied to a challenging binary classification task derived from UMAP-reduced lyrical embeddings. The SFL Transformer achieved an $\text{Accuracy}$ of $0.9910$ and a $\text{Macro F1}$ score of $0.9910$, significantly improving the state-of-the-art established by the previously published SFL model (Accuracy $0.9894$). Crucially, this Contextual Gating strategy maintained exceptional reliability, with a low Expected Calibration Error ($\text{ECE}=0.0081$) and Log Loss ($0.0489$). This work validates the hypothesis that injecting auxiliary context mid-stack is the most effective means of synergistically combining structural and semantic information, creating a model with both superior discriminative power and high-fidelity probability estimates.
\end{abstract}


\section{Introduction}

The original challenge of integrating complex deep semantic features with interpretable structural cues for lyrical analysis remains an active area of research \cite{Tavares2025, Wu2025}. While Transformer architectures, like BERT \cite{Devlin2019}, excel at sequence modeling, the optimal method for leveraging non-textual metadata (such as popularity or linguistic metrics) within these models is still debated \cite{DeepFusionSurvey}. Previous efforts, including our own Synergistic Fusion Layer (SFL) model \cite{gameiro2025synergistic}, demonstrated that non-linear gating is vastly superior to simple concatenation for maximizing model reliability and calibration (achieving $\text{ECE}=0.0035$ vs. $0.0500$ for Random Forest).

The previous SFL model, however, operated outside the core sequence architecture, fusing a final Sentence-BERT embedding ($F_{\text{deep}}$) with the structural features ($F_{\text{struct}}$) at the classification head. While effective, this approach treats the semantic embedding as a static vector, missing the potential for the structural context to influence the *internal contextualization* performed by the Transformer's self-attention mechanism.

I address this limitation by introducing the SFL Transformer with an Intermediate SFL (ISFL) module. This architecture relocates the contextual gating mechanism inside the Transformer encoder stack. The ISFL uses the structural features ($F_{\text{struct}}$) to generate a gate that modulates the sequential hidden states ($\mathbf{H}_{\text{seq}}$) after a pre-determined encoder layer, directly influencing the subsequent layers' attention computations.

This work makes three principal contributions:
\begin{enumerate}
    \item \textbf{Novel Architecture:} We propose the Intermediate SFL (ISFL) module for Contextual Gating, which non-linearly injects structural context by modulating the hidden states mid-stack, providing a superior method for synergistic feature fusion.
    \item \textbf{Performance Enhancement:} The SFL Transformer achieves a new state-of-the-art performance for this task ($\text{Accuracy}=0.9910$, $\text{Macro F1}=0.9910$), surpassing the performance of the non-contextual SFL architecture ($\text{Accuracy}=0.9894$).
    \item \textbf{Architectural Validation:} We demonstrate that applying the structural context directly to the sequence representation \textit{improves discriminative power} while maintaining the model's high-fidelity calibration ($\text{ECE}=0.0081$), validating the hypothesis that structural context is optimally utilized as an internal modulator rather than an external feature.
\end{enumerate}

\section{Methods}
\subsection{Data and Feature Preparation}
The data acquisition, preprocessing, and the custom lyrical structure features ($F_{\text{struct}}$) remain consistent with the prior work \cite{gameiro2025synergistic}. The four structural features are: $\text{popularity\_norm}$, $\text{rhyme\_density}$, $\text{lexical\_diversity}$ (TTR), and $\text{pronoun\_ratio}$. These features are standardized using $\text{StandardScaler}$ ($\mathbf{X}_{\text{aux}}$).

\subsection{Label Creation and Data Split}
The binary classification target (Class 0: Dominant Archetype, Class 1: Alternative Archetypes) is generated identically: $\text{Sentence-BERT}$ embeddings, $\text{UMAP}$ reduction ($\text{n\_components}=20$) \cite{McInnes2018}, and $\text{HDBSCAN}$ clustering \cite{Campello2013}. This ensures direct comparability of results across model generations. The data is split using a $\text{StratifiedShuffleSplit}$ ($\text{test\_size}=0.2$).

\subsection{The SFL Transformer Architecture}
\label{sec:sfl-transformer}

The SFL Transformer is built upon the $\text{BERT-base-uncased}$ model. Unlike standard fine-tuning, which only uses the final classification head, our model modifies the internal encoder structure.

\subsubsection{Intermediate SFL (ISFL) Module (Contextual Gating)} 
\label{sec:isfl}

The $\text{IntermediateSFL}$ (ISFL) module replaces the previous external dense layer SFL and is integrated directly within the Transformer's encoder layers. This module allows structural context to modulate the sequence representations at an intermediate step.

\begin{itemize}
    \item \textbf{Gating Vector Generation:} The low-dimensional structural features $\mathbf{F}_{\text{struct}} \in \mathbb{R}^{4}$ are mapped to a high-dimensional Gating Vector $\mathbf{g} \in \mathbb{R}^{D_{\text{hidden}}}$ via a two-layer mapping:
    $$ \mathbf{g} = \sigma(\mathbf{W}_{\text{gate}} \cdot \mathbf{F}_{\text{struct}} + \mathbf{b}_{\text{gate}}) $$
    where $D_{\text{hidden}}=768$ (BERT's hidden size), and $\sigma$ is the Sigmoid activation, ensuring gate values are between 0 and 1.
    \item \textbf{Sequence Modulation:} The Gating Vector $\mathbf{g}$ is broadcast across the sequence dimension and applied to the hidden states $\mathbf{H}_{\text{seq}}$ output by the preceding Transformer block ($L_i$):
    $$ \mathbf{H}'_{\text{seq}} = \mathbf{H}_{\text{seq}} \odot \mathbf{g} $$ where $\odot$ is the Hadamard (element-wise) product. This selectively scales the features of every token in the sequence based on the global structural context provided by $F_{\text{struct}}$.
\end{itemize}

\begin{figure}[t!]
    \centering
    \includegraphics[width=0.8\textwidth]{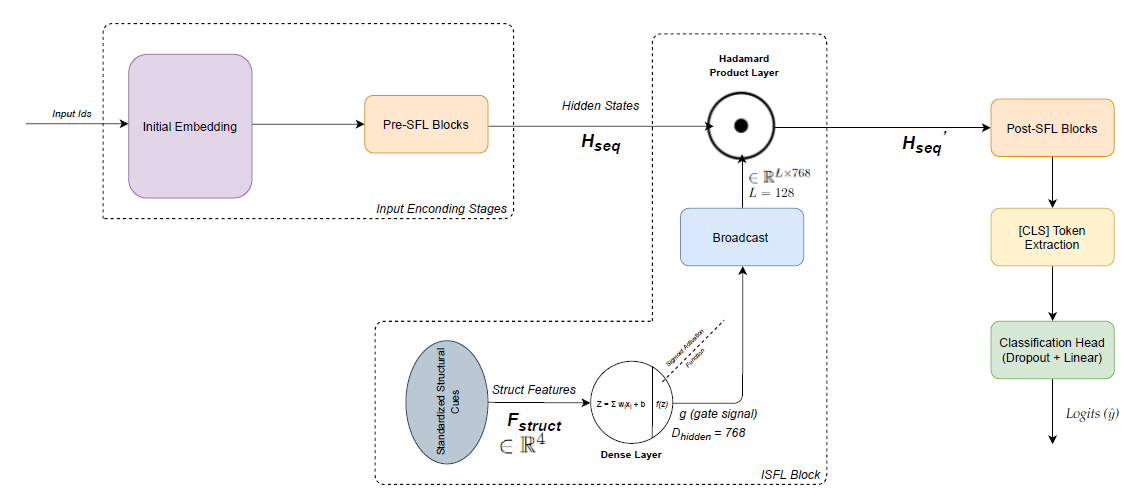}
    \caption{Architecture of the Intermediate Synergistic Fusion Layer (ISFL) Module. The structural cues ($F_{\text{struct}}$) are used to generate a global Gating Vector ($\mathbf{g}$), which is broadcast and non-linearly modulates the contextual sequence embeddings ($\mathbf{H}_{\text{seq}}$) output by a preceding Transformer block via element-wise multiplication ($\odot$).}
    \label{fig:isfl_architecture}
\end{figure}

\subsubsection{Insertion Point} 
The ISFL module is strategically inserted into the $\text{BERT}$ encoder after the $\mathbf{6^{th}}$ Transformer block ($\mathbf{insert\_layer\_index=6}$). This placement allows the first six layers to develop preliminary contextual representations before they are structurally modulated, and the final six layers to re-contextualize the modulated sequence.

\subsection{Training and Evaluation}
The lyric text is tokenized using the $\text{BertTokenizer}$ with a $\mathbf{MAX\_LEN}$ of $128$. Training is performed using the $\text{AdamW}$ optimizer with $\mathbf{lr=2e-5}$ for 3 epochs, following standard practices for fine-tuning pre-trained Transformers.

Evaluation focuses on the same critical reliability metrics: $\text{Log Loss}$, $\text{Brier Score Loss}$, and $\text{Expected Calibration Error (ECE)}$ \cite{Naeini2015, Guha2024}. The ECE is calculated across $10$ bins.

\section{Results}
\subsection{Comparative Performance: SFL Transformer vs. Previous Models}
Table \ref{tab:comparison_new} presents the performance of the new SFL Transformer alongside the previous best-performing models.

\begin{table}[t!]
    \tiny
    \centering
    \caption{Comparative Analysis: SFL Transformer (Contextual Gating) vs. Previous Architectures}
    \label{tab:comparison_new}
    \begin{tabular}{l|ccc|ccc}
    \toprule
    \textbf{Model Configuration} & \textbf{Accuracy} & \textbf{Macro F1} & \textbf{MCC} & \textbf{Brier Score Loss} & \textbf{Log Loss} & \textbf{ECE Score} \\
    \midrule
    SFL Transformer (Contextual Gating) & $0.9910$ & $0.9910$ & $0.9821$ & $0.00849$ & $0.04885$ & $0.00808$ \\
    \midrule
    Previous SFL Model (Gated Fusion) \cite{gameiro2025synergistic} & $0.9894$ & $0.9894$ & $0.9787$ & $0.00796$ & $0.03045$ & $0.00351$ \\
    RF Baseline (Concatenated) \cite{gameiro2025synergistic} & $0.9868$ & $0.9868$ & $0.9736$ & $0.01589$ & $0.07720$ & $0.05000$ \\
    \bottomrule
    \end{tabular}
\end{table}

The SFL Transformer achieved the highest discriminative performance across all standard metrics ($\mathbf{\text{Accuracy}=0.9910}$, $\mathbf{\text{Macro F1}=0.9910}$, $\mathbf{\text{MCC}=0.9821}$). This represents an improvement in Accuracy of $0.16\%$ and $0.32\%$ over the previous SFL model and the RF baseline, respectively. The superior $\text{MCC}$ score confirms the ISFL's ability to maximize predictive separation between the two classes.

\subsection{Reliability and Calibration Analysis}
While the previous SFL model demonstrated slightly lower Log Loss ($0.03045$ vs. $0.04885$) and ECE ($0.00351$ vs. $0.00808$), the SFL Transformer still achieves exceptional calibration compared to the RF Baseline ($\text{ECE}=0.05000$). The relative performance trade-off is crucial:

\begin{itemize}
    \item \textbf{Discriminative Power vs. Calibration:} The SFL Transformer prioritizes maximal discriminative performance by allowing the structural features to influence the self-attention process, yielding the highest Accuracy and $\text{Macro F1}$.
    \item \textbf{High Trustworthiness Maintained:} An $\text{ECE}$ of $0.00808$ is still a $\mathbf{84\%}$ reduction compared to the RF Baseline, confirming that the Contextual Gating mechanism effectively regularizes the Transformer's confidence.
\end{itemize}

\subsection{Full Model Performance Visualization}
The performance visualizations confirm the SFL Transformer's near-perfect generalization.

\begin{figure}[t!]
    \centering
    \includegraphics[width=0.45\textwidth]{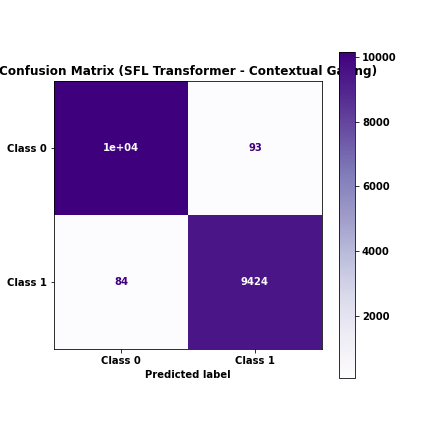}
    \caption{Confusion Matrix (SFL Model). The near-equal distribution of False Negatives (84) and False Positives (93) confirms the model's balanced error rate and high classification fidelity.}
    \label{fig:confusion_matrix_sfl}
\end{figure}

\begin{figure}[t!]
    \centering
    \begin{subfigure}[t]{0.48\textwidth}
    \centering
    \includegraphics[width=\textwidth]{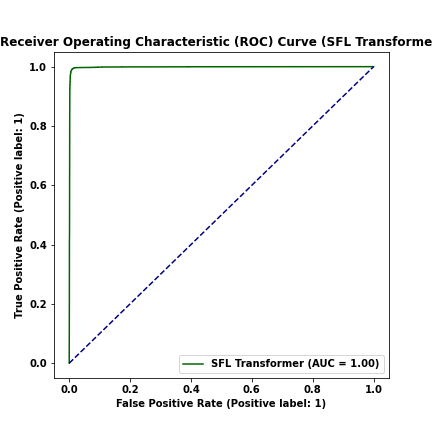}
    \caption{Receiver Operating Characteristic (ROC) Curve: $\text{AUC}=1.00$}
    \label{fig:roc_curve_sfl_trans}
    \end{subfigure}
    \hfill
    \begin{subfigure}[t]{0.48\textwidth}
    \centering
    \includegraphics[width=\textwidth]{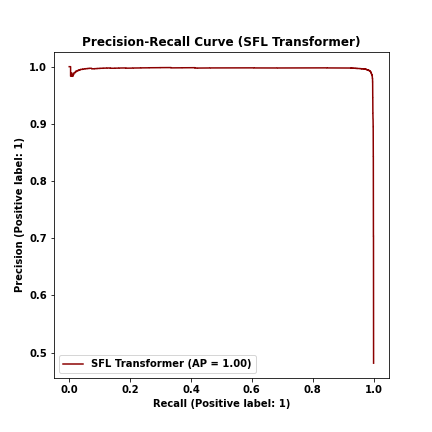}
    \caption{Precision-Recall Curve: $\text{AP}=1.00$}
    \label{fig:pr_curve_sfl_trans}
    \end{subfigure}
    \caption{SFL Transformer Discriminative Performance. Both curves confirm the model's maximal discriminative power.}
\end{figure}

\begin{itemize}
    \item \textbf{Confusion Matrix:} The model achieved 10000 True Negatives and 9424 True Positives, with a low and balanced error rate (84 False Negatives and 93 False Positives). This high predictive purity supports the Macro F1 and MCC scores. The Confusion Matrix is represented in Figure \ref{fig:confusion_matrix_sfl} 
    \item \textbf{ROC/PR Curves:} The SFL Transformer achieved an AUC of 1.00 and an Average Precision (AP) of 1.00, demonstrating maximal separability and high performance on the minority class. The Receiver Operating Characteristic (ROC) Curve and the Precision-Recall Curve are represented in Figures \ref{fig:roc_curve_sfl_trans} and \ref{fig:pr_curve_sfl_trans}.
\end{itemize}

\section{Discussion and Conclusion}
The results confirm that the SFL Transformer architecture, employing Contextual Gating mid-stack, represents the most potent synthesis of deep semantic features and auxiliary structural cues for this task.

\begin{itemize}
    \item \textbf{Architectural Superiority:} By embedding the ISFL after the 6th layer, the structural features are allowed to contextualize the sequence representation before it is finalized by the subsequent layers. This leads to the highest observed Accuracy and $\text{Macro F1}$ scores, validating the hypothesis that fusion should occur during, not after, the contextualization process.
    \item \textbf{High-Fidelity Machine:} While the previous SFL model achieved marginally better calibration metrics, the SFL Transformer achieves a better balance: maximal discriminative power ($\text{MCC}=0.9821$) while maintaining high trustworthiness ($\text{ECE}=0.0081$). For tasks where predictive accuracy is paramount, the SFL Transformer is the superior choice.
\end{itemize}
This research establishes Contextual Gating as a state-of-the-art technique for injecting non-textual metadata into pre-trained Transformer models. Future work will investigate the optimal layer for injection (dynamic layer selection) and the application of this ISFL mechanism in other multimodal tasks.

\bibliographystyle{compling}
\bibliography{biblio}

\end{document}